\documentclass[fleqn,twoside]{article}
\usepackage[headings]{espcrc2}

\readRCS
$Id: espcrc2.tex,v 1.2 2004/02/24 11:22:11 spepping Exp $
\usepackage{graphicx, balance}
\usepackage[figuresright]{rotating}
\usepackage{epstopdf}

\newcommand{\AmS}{{\protect\the\textfont2
  A\kern-.1667em\lower.5ex\hbox{M}\kern-.125emS}}

\title{\textbf{Enhancing Template Security of Face Biometrics by Using Edge Detection and Hashing}}

\author{Manoj Krishnaswamy\address[DCSE]{Research Scholar, Department of Studies in Computer Science, University of Mysore, Mysore, Contact: manojkrishnaswamy@gmail.com \\},
G. Hemantha Kumar\address{Professor, Department of Studies in Computer Science, University of Mysore, Mysore.}}

\runtitle{Enhancing Template Security of Face Biometrics by Using Edge Detection and Hashing}
\runauthor{Manoj Krishnaswamy, et al.,}

\begin{document}
\begin{abstract}
In this paper we address the issues of using edge detection techniques on facial images to produce cancellable biometric templates and a novel method for template verification against tampering. With increasing use of biometrics, there is a real threat for the conventional systems using face databases, which store images of users in raw and unaltered form. If compromised not only it is irrevocable, but can be misused for cross-matching across different databases. So it is desirable to generate and store revocable templates for the same user in different applications to prevent cross-matching and to enhance security, while maintaining privacy and ethics. By comparing different edge detection methods it has been observed that the edge detection based on the Roberts Cross operator performs consistently well across multiple face datasets, in which the face images have been taken under a variety of conditions. And we have proposed a novel scheme using hashing, for extra verification, in order to harden the security of the stored biometric templates.  \\\\
{\bf Keywords :} Cancellable Biometrics, Edge Detection, Face Biometrics, Template Security.
\end{abstract}

\maketitle

\section{INTRODUCTION}
The dimensions, proportions and physical attributes of a person's face are 
unique. Biometric facial recognition systems will measure and analyze the 
overall structure, shape and proportions of the face: Distance between the 
eyes, nose, mouth, and jaw edges; upper outlines of the eye sockets, the 
sides of the mouth, the location of the nose and eyes, the area surrounding 
the cheekbones. At enrolment, several pictures are taken of the user's face, 
with slightly different angles and facial expressions, to allow for more 
accurate matching. For verification and identification, the user stands in 
front of the camera for a few seconds, and the scan is compared with the 
template previously recorded. Benefits of face biometric systems being that 
it is not intrusive, can be done from a distance, even without the user 
being aware of it (for instance when scanning the entrance to a bank or a 
high security area). Weaknesses of face biometric systems: Face biometric 
systems are more suited for authentication than for identification purposes, 
as it is easy to change the proportion of one's face by wearing a mask, a 
nose extension, etc. Also, user perceptions / civil liberty: Most people are 
uncomfortable with having their picture taken. Applications of face 
biometrics include access to restricted areas and buildings, banks, 
embassies, military sites, airports, law enforcement.

One advantage of passwords over biometrics is that they can be re-issued. If 
a token or a password is lost or stolen, it can be cancelled and replaced by 
a newer version. This is not naturally available in biometrics. If someone's 
face is compromised from a database, they cannot cancel or reissue it. 
Cancellable biometrics is a way in which to incorporate protection and the 
replacement features into biometrics. It was first proposed by N. K. Ratha, 
J. H. Connell and R. M. Bolle \cite{1}.

Several methods for generating cancellable biometrics have been proposed. 
The first fingerprint based cancellable biometric system was designed and 
developed by S. Tulyakov, F. Farooq and V. Govindaraju \cite{2}. Essentially, 
cancellable biometrics performs a distortion of the biometric image or 
features before matching. The variability in the distortion parameters 
provides the cancellable nature of the scheme. Some of the proposed 
techniques operate using their own recognition engines, such as A. B. J. 
Teoh, A. Goh and D. C. L. Ngo \cite{3} and M. Savvides, B. V. K. V. Kumar and P. 
K. Khosla \cite{4}. Whereas other methods, such as M. A. Dabbah, W. L. Woo and S. 
S. Dlay \cite{5} take the advantage of the advancement of the well-established 
biometric research for their recognition front-end to conduct recognition. 
Although this increases the restrictions on the protection system, it makes 
the cancellable templates more accessible for available biometric 
technologies.

\subsection{Edge Detection}
The result of applying an edge detector to an image may lead to a set of 
connected curves that indicate the boundaries of objects, the boundaries of 
surface markings as well as curves that correspond to discontinuities in 
surface orientation. Thus, applying an edge detection algorithm to an image 
may significantly reduce the amount of data to be processed and may 
therefore filter out information that may be regarded as less relevant, 
while preserving the important structural properties of an image. The edge 
detection filters used for experimentation are based on Discrete Laplace 
operator, Sobel operator \cite{6}, Roberts Cross operator \cite{7}, Frei-Chen operator \cite{8} and Prewitt operator \cite{9}.

\subsection{Cryptosystems in Biometrics}
Biometric cryptosystems as discussed by Y. Dodis, R. Ostrovsky, L. Reyzin 
and A. Smith \cite{10}, F. Hao, R. Anderson and J. Daugman \cite{11}, K. Nandakumar, 
A.K. Jain and S. Pankanti \cite{12}, Y. Sutcu, Q. Li and N. Memon \cite{13}, use 
techniques that associate an external key with a user's biometric to obtain 
helper data. The helper data should not reveal any significant information 
about the template or the key and at the same time it can be used to recover 
the key when the original biometric is presented. The concept of data hiding 
in digital watermarks has been discussed by C. T. Hsu and J. L. Wu \cite{14}. 
Encryption algorithm to secure the image using fingerprint and password has 
been discussed by Manvjeet Kaur, Dr. Sanjeev Sofat and Deepak Saraswat \cite{15} 
involves more time consuming methods.

\subsection{Secure Hash Algorithm}
A hash function is an algorithm that transforms (hashes) an arbitrary 
set of data elements, such as a text file, into a single fixed length value (the hash). The computed hash value may then be used to verify the integrity of copies of the original data without providing any means to derive said original data. This irreversibility means that a hash value may be freely distributed or stored, as it is used for comparative purposes only. SHA stands for Secure Hash Algorithm. SHA-2 includes a significant number of changes from its predecessor, SHA-1. SHA-2 consists of a set of four hash functions with digests that are 224, 256, 384 or 512 bits. We have used SHA-256 for our experimentation.\\
The security provided by a hashing algorithm is entirely dependent upon its ability to produce a unique value for any specific set of data. When a hash function produces the same hash value for two different sets of data then a collision is said to occur. Collision raises the possibility that an attacker may be able to computationally craft sets of data which provide access to information secured by the hashed values of pass codes or to alter computer data files in a fashion that would not change the resulting hash value and would thereby escape detection. A strong hash function is one that is resistant to such computational attacks. A weak hash function is one where a computational approach to producing collisions is believed to be possible. A broken hash function is one where a computational method for producing collisions is known to exist. 
In 2005, security flaws were identified in SHA-1, namely that a mathematical weakness might exist, indicating that a stronger hash function would be desirable. Although SHA-2 bears some similarity to the SHA-1 algorithm, these attacks have not been successfully extended to SHA-2.

\subsection{Advanced Encryption Standard}
The Advanced Encryption Standard (AES) is a specification for the encryption of electronic data established by the U.S. National Institute of Standards and Technology (NIST) in 2001 \cite{20}. The algorithm described by AES is a symmetric-key algorithm, meaning the same key is used for both encrypting and decrypting the data. AES is based on a design principle known as a substitution-permutation network, and is fast in both software and hardware. We have used AES to generate key size of 256 bits (AES-256). High speed and low RAM requirements were criteria of the AES selection process. Thus AES performs well on a wide variety of hardware, from 8-bit smart cards to high-performance computers.

\section{PROPOSED METHOD}
\subsection{Enrolment (shown in Figure 1)}
\begin{figure*}
\centering
\includegraphics[width=5.00in,height=2.00in]{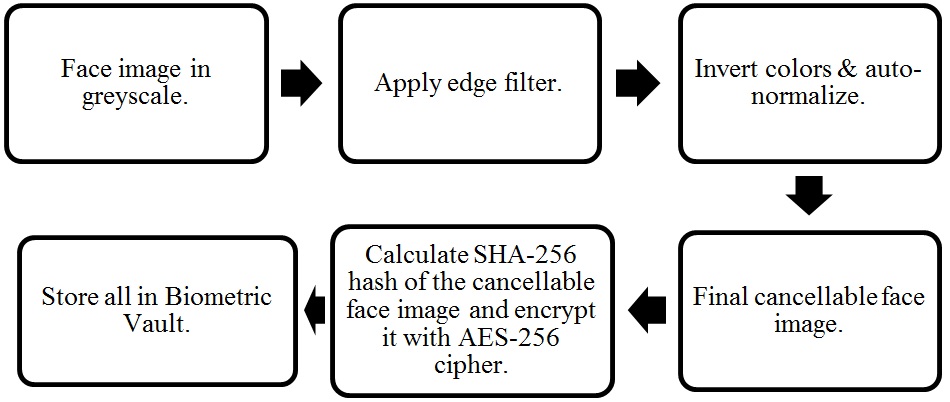}
\caption{Block diagram of proposed method for enrolment.}
\end{figure*}

\textit{Step 1:} Input face image from dataset (ATT, YALE or IFD) which is in greyscale (or 
converted). The ATT dataset of faces (formerly 'The ORL Database of Faces'), 
YALE dataset and IFD (Indian Face Dataset) are unmodified except for 
conversion to JPEG and renaming of the files. Datasets used are ATT, IFD and YALE with sample images shown in Figure 3, Figure 4 and Figure 5 respectively.

\textit{Step 2:} Apply edge detection filter. The edge detection filters used for 
experimentation are based on Discrete Laplace operator, Sobel operator, 
Roberts Cross operator, Frei-Chen operator and Prewitt operator.

\textit{Step 3:} Invert colors of the image and then auto normalize. This is done because a 
major drawback to application of the edge detection filters is an inherent 
reduction in overall image contrast produced by the operation, which is in 
turn used to become an advantage in our case since it provides obscuring the 
original image to an acceptable level. Normalize stretches the histogram, so 
the whole range of colors is used as to get more information out of the 
image. Hence by inverting the filtered image and auto normalizing we get the 
contrast to an acceptable level.

\textit{Step 4:} Calculate SHA-256 hash value of the final cancellable face image and encrypt it with AES-256 bit cipher. The AES-256 bit cipher is a symmetric key algorithm which uses the same password for encrypting and decrypting.

\textit{Step 5:} Obtain the final filtered face image and store in the corresponding 
dataset. Dataset ATT-L is the set obtained after applying step 2 with 
Laplace edge detector filter and step 3 on ATT dataset (sample shown in Figure 6.
Dataset ATT-S is the set obtained after applying step 2 with Sobel edge 
detector filter and step 3 on ATT dataset (sample shown in Figure 7). Dataset 
ATT-R is the set obtained after applying step 2 with Roberts edge detector 
filter and step 3 on ATT dataset (sample shown in Figure 8). Dataset ATT-F is 
the set obtained after applying step 2 with Frei-Chen edge detector filter 
and step 3 on ATT dataset (sample shown in Figure 9). Dataset ATT-P is the set 
obtained after applying step 2 with Prewitt edge detector filter and step 3 
on ATT dataset (sample shown in Figure 10).Similarly for YALE and IFD datasets.

\subsection{Verification (shown in Figure 2)}

\begin{figure*}
\centering
\includegraphics[width=5.00in,height=2.00in]{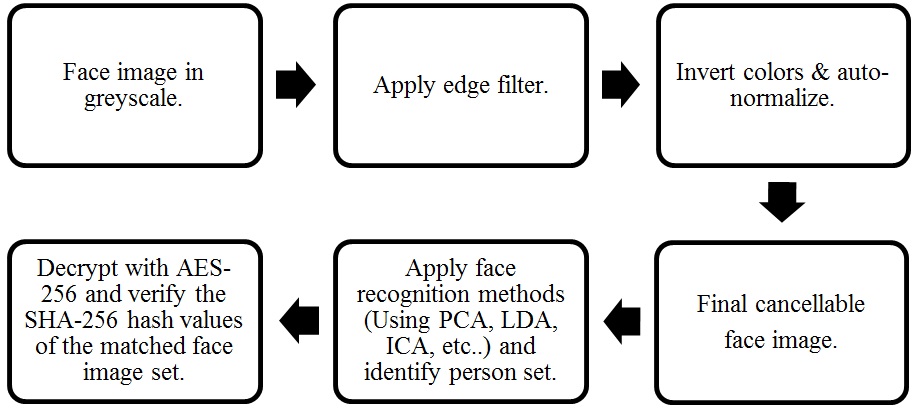}
\caption{Block diagram of proposed method for verification.}
\end{figure*}

\textit{Steps 1 to 3:} Same as in Enrolment.

\textit{Step 4:} Apply face recognition methods to identify the person set. In our experiment the face recognition methods use the following: PCA (Principal Component Analysis), IPCA (Incremental PCA), LDA (Linear Discriminant Analysis) and ICA (Independent Component Analysis).

\textit{Step 5:} For the set of images of the matched person, verify the SHA-256 hash values after decrypting with AES-256 cipher. This step ensures that the stored biometric templates have not been tampered with.

The proposed method starts with a non-invertible feature transformation by using edge detection filters and is combined with a key binding biometric crypto system. The SHA-256 hash value, which is AES-256 bit encrypted, helps in binding the cancellable template with an encrypted key. The AES-256 bit cipher is a symmetric algorithm and hence uses the same password for encryption and decryption. Here the password that is used to bind the values can be user driven or be at the system level, depending on the feasibility of the biometric system.

Our proposed method focuses on the Roberts' Cross based edge detector due to 
its consistently highest matching accuracy across different datasets [Table 
1, Table 2, and Table 3]. According to Roberts, an edge detector should have 
the following properties: the produced edges should be well-defined, the 
background should contribute as little noise as possible, and the intensity 
of edges should correspond as close as possible to what a human would 
perceive.

After applying the edge filter(s), the image colors are inverted since edge 
filters discard other information than the detected edges (first image of 
Figure 11). The image is then auto normalized edges (second image of Figure 11) 
to the full dynamic range, to further enhance the remaining details. In 
image processing, normalization is a process that changes the range of pixel 
intensity values. Applications include photographs with poor contrast due to 
glare, for example. Normalization is sometimes called contrast stretching. 
In more general fields of data processing, such as digital signal 
processing, it is referred to as dynamic range expansion. The purpose of 
dynamic range expansion in the various applications is usually to bring the 
image, or other type of signal, into a range that is more familiar or normal 
to the senses, hence the term normalization. Often, the motivation is to 
achieve consistency in dynamic range for a set of data, signals, or images 
to avoid mental distraction or fatigue. For example, a newspaper will strive 
to make all of the images in an issue share a similar range of greyscale.

Normalization is a linear process. If the intensity range of the image is 50 
to 180 and the desired range is 0 to 255 the process entails subtracting 50 
from each of pixel intensity, making the range 0 to 130. Then for each pixel 
the intensity is multiplied by 255/130, making the range 0 to 255. 
Auto-normalization in image processing software typically normalizes to the 
full dynamic range of the number system specified in the image file format. 
We have found through our experiment that when the face images are auto 
normalized after applying edge filter and inverting colors, matching accuracy 
increases.

\begin{figure*}
\centering
\parbox{6.5cm}{
\includegraphics[height=2.3cm]{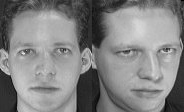}
\caption{From ATT dataset.}}
\end{figure*}

\begin{figure*}
\centering
\parbox{6.5cm}{
\includegraphics[height=2.3cm]{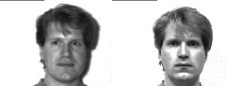}
\caption{From Yale dataset.}}
\end{figure*}

\begin{figure*}
\centering
\parbox{6.5cm}{
\includegraphics[height=2.3cm]{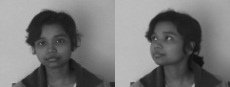}
\caption{From IFD dataset.}}
\end{figure*}

\begin{figure*}
\centering
\parbox{6.5cm}{
\includegraphics[height=2.3cm]{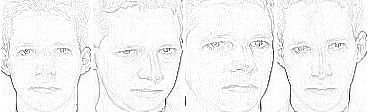}
\caption{From ATT-L.}}\qquad
\end{figure*}

\begin{figure*}
\centering
\parbox{6.5cm}{
\includegraphics[height=2.3cm]{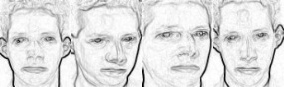}
\caption{From ATT-S.}}
\end{figure*}

\begin{figure*}
\centering
\parbox{6.5cm}{
\includegraphics[height=2.3cm]{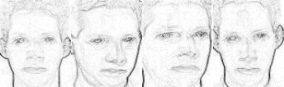}
\caption{From ATT-R.}}\qquad
\end{figure*}

\begin{figure*}
\centering
\parbox{6.5cm}{
\includegraphics[height=2.3cm]{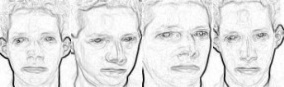}
\caption{From ATT-F.}}
\end{figure*}

\begin{figure*}
\centering
\parbox{6.5cm}{
\includegraphics[height=2.3cm]{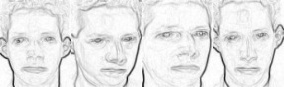}
\caption{From ATT-P.}}\qquad
\end{figure*}

\begin{figure*}
\centering
\parbox{6.5cm}{
\includegraphics[height=2.8cm]{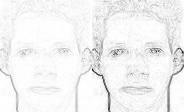}
\caption{Before and after auto normalization.}}
\end{figure*}

\section{RESULTS AND DISCUSSIONS}
\begin{table*}
\centering
\caption{\bf Facial recognition accuracy of ATT dataset and variants}
\begin{tabular}{|p{45pt}|p{42pt}|p{42pt}|p{42pt}|p{42pt}|p{42pt}|p{42pt}|}
\hline
\multicolumn{7}{|p{300pt}|}{Accuracy ({\%}) (Recognition Rate)}  \\
\hline
\raisebox{-1.50ex}[0cm][0cm]{Classifier}& 
\multicolumn{6}{|p{300pt}|}{Dataset}  \\
\cline{2-7} 
 & 
ATT& 
ATT-L& 
ATT-S& 
ATT-R& 
ATT-F& 
ATT-P \\
\hline
ICA& 
91.3& 
83.1& 
86.9& 
86.9& 
87.5& 
88.8 \\
\hline
IPCA& 
93.1& 
87.5& 
89.4& 
88.1& 
88.1& 
90.0 \\
\hline
LDA& 
\textbf{94.4}& 
\textbf{91.3}& 
\textbf{93.1}& 
\textbf{93.1}& 
\textbf{92.5}& 
\textbf{91.9} \\
\hline
PDA& 
91.3& 
83.1& 
88.1& 
87.5& 
88.8& 
88.1 \\
\hline
\end{tabular}
\end{table*}

The ATT dataset \cite{17} comprises of face frontal images with low resolution 
(92x112 pixels). The images have dark background (Figure 3) with most of it 
not present in the images by comparison to other datasets. From Table 1, LDA 
based face recognition method is having the best matching accuracy. The 
proposed method with Roberts Cross filter and Sobel filter are showing the 
least variation (1.3{\%}) w.r.t. matching accuracy.

\begin{table*}
\centering
\caption{\bf Recognition accuracy of face recognition methods of YALE dataset and variants}
\begin{tabular}{|p{45pt}|p{42pt}|p{42pt}|p{42pt}|p{42pt}|p{42pt}|p{42pt}|}
\hline
\multicolumn{7}{|p{300pt}|}{Accuracy ({\%}) (Recognition Rate)}  \\
\hline
\raisebox{-1.50ex}[0cm][0cm]{Classifier}& 
\multicolumn{6}{|p{292pt}|}{Dataset}  \\
\cline{2-7} 
 & 
YALE& 
YALE-L& 
YALE-S& 
YALE-R& 
YALE-F& 
YALE-P \\
\hline
ICA& 
83.3& 
85.0& 
81.7& 
78.3& 
86.7& 
81.7 \\
\hline
IPCA& 
71.7& 
68.3& 
75.0& 
73.3& 
78.3& 
75.0 \\
\hline
LDA& 
\textbf{91.7}& 
\textbf{86.7}& 
\textbf{88.3}& 
\textbf{90.0} & 
\textbf{90.0}& 
\textbf{90.0} \\
\hline
PCA& 
81.7& 
86.7& 
83.3& 
80.0& 
85.0& 
86.7 \\
\hline
\end{tabular}
\end{table*}

The YALE dataset \cite{18} comprises of face frontal images with medium 
resolution (320x243 pixels). The face images (Figure 4) have mostly a bright 
background and a few with shadows. Those shadows are subsequently removed 
due to edge detection filtering. From Table 2, LDA based face recognition 
method is having the best matching accuracy. The proposed method with 
Roberts Cross filter, Frei-Chen filter and Prewitt filter are showing the 
least variation (1.7{\%}) w.r.t. matching accuracy.

\begin{table*}
\centering
\caption{\bf Recognition accuracy of face recognition methods of IFD dataset and variants}
\begin{tabular}{|p{45pt}|p{42pt}|p{42pt}|p{42pt}|p{42pt}|p{42pt}|p{42pt}|}
\hline
\multicolumn{7}{|p{300pt}|}{Accuracy ({\%}) (Recognition Rate)}  \\
\hline
\raisebox{-1.50ex}[0cm][0cm]{Classifier}& 
\multicolumn{6}{|p{292pt}|}{Dataset}  \\
\cline{2-7} 
 & 
IFD& 
IFD-L& 
IFD-S& 
IFD-R& 
IFD-F& 
IFD-P \\
\hline
ICA& 
76.7& 
76.3& 
75.4& 
78.0& 
75.0& 
75.0 \\
\hline
IPCA& 
76.7& 
86.0& 
86.9& 
86.0& 
86.0& 
86.4 \\
\hline
LDA& 
\textbf{91.1}& 
\textbf{89.8}& 
\textbf{89.8}& 
\textbf{89.8} & 
\textbf{89.4}& 
\textbf{89.0} \\
\hline
PCA& 
76.3& 
72.0& 
76.3& 
75.8& 
76.7& 
77.1 \\
\hline
\end{tabular}
\end{table*}

The IFD dataset \cite{19} comprises of face frontal and some side poses images 
with high resolution (640x480 pixels). The face images (Figure 5) have mostly 
a dull background. From Table 3, LDA based face recognition method is having 
the best matching accuracy. The proposed method with Roberts Cross filter, 
Laplace filter and Sobel filter are showing the least variation (1.3{\%}) 
w.r.t. matching accuracy.

The facial images across all the datasets used here are taken under 
controlled conditions and are less susceptible to noise. After applying the 
edge filter and inverting colors, we have further enhanced the image by auto 
normalization. The face recognition method which uses LDA combined with 
Roberts Cross filter in our proposed scheme shows the highest matching 
accuracy consistently across the wide range of facial images of different 
types of datasets used. There is insignificant changes w.r.t. matching 
accuracy (varying from 1.3{\%} to 1.7{\%}) across the datasets, with and 
without our proposed method.

Template security emphasizes on obscuring the template images, the slight 
reduction in accuracy is definitely acceptable. Robert's filter is 
mathematically the simplest of all the compared edge detection methods. 
Hence, the proposed scheme has low impact on the speed of execution, hence 
can be incorporated into existing systems without too much overhead. The 
proposed scheme can be incorporated at the time of enrolment and 
verification itself. The filtered images can thus be stored instead of the 
unaltered face images, thereby providing a form of encryption. Since the 
subsequent images taken for enrolment are binary different even when it is 
with same camera and lighting conditions, the obscured template will be 
different. This in turn provides a non-invertible template, where in case 
the dataset of the filtered images (used for matching) is compromised, it 
can be revoked a new set generated without worrying about misuse of the lost 
data. Also, by varying the convolution kernel values of the Robert's filter 
gradient, more cancellable templates can be generated for a particular face 
image, as discussed for difference of gaussian edge filter by G. Hemantha 
Kumar and Manoj Krishnaswamy \cite{16}.

\begin{table*}
\centering
\caption{\bf SHA-256 Hash of cancellable image before and after AES-256 bit cipher}
\begin{tabular}{|p{150pt}|p{250pt}|}
\hline
 & Cancellable template \\
\hline
SHA-256 Hash unencrypted in HEX representation: & 
5BB015B9A86F88AA1C21C47170553C3A3FB0052F35FF
B35993D3AA0C43988449 \\
\hline
SHA-256 Hash encrypted with AES-256 cipher (key=1234) in HEX 
representation: & 
5A8B2D5EDB8D5AEED90F67F2D7868C62DCD1B81E0
D588FB2A00111ABF5736589E649E6514AC256B74532E2
6AE5D369CBFF3715845E7B91C7223A877591082051FF2
94EBA0B0B7632C3C6BE5936A2078DA86487D39CDB2D
B41A43FB53A2330F85021AEA394F3E867155979CF3BE7
A037209CDAC7E2D3896A200C89903EE48F36AFAA0F14
9F8A1D07C871537FB86EDB4AC 
\\
\hline
\end{tabular}
\end{table*}

By storing SHA-256 hash of the stored biometric template and encrypting with AES-256 algorithm (Table 4) we have provided a strong measure against biometric template tampering. SHA-256 hashing and AES-256 cipher can be performed computationally fast (less than a second) and hence can be easily incorporated into existing systems. Although we have assumed that the attacker will not be able to easily gain access on the various levels to compromise the entire system, even in case the entire system was being compromised, the cancellable templates can be re-issued which provides new hash values automatically. Due to the non-invertible nature of the templates there is no worry of misuse of lost data. Other schemes involve calculating the helper data (in our case the ciphered hash value) for each set of biometric templates which becomes time consuming during verification. The time taken to decrypt only once enhances the speed of execution and can be incorporated in systems which require speed as well as security. Useful scenarios for the proposed method could be in real time systems, banking, ATM access, etc.

\section {CONCLUSIONS}
We have shown that the final filtered images itself can be used for face matching instead of unaltered face images. The results are checked across datasets which encompasses a wide variety of images taken under different conditions as well as different resolutions and image quality. We proposed a novel method for generating cancellable face biometrics and to secure the stored templates in a way which is suitable for integration with current face matching systems with acceptable alterations.\\
Also, by using fast, proven and standard hashing (SHA-256) and cryptographic (AES-256) methods  for data verification, the vault is further enhanced. We discussed their strengths and shortcomings, as well as their relative performance on different databases under a variety of conditions. The approach allows for enhanced template security, privacy and maintaining good ethics in biometric systems. It is important that biometrics based authentication systems are designed to withstand different sources of attacks on the system.

\noindent{\includegraphics[width=1in,height=1.7in,clip,keepaspectratio]{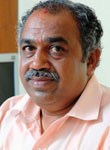}}
\begin{minipage}[b][1in][c]{2.0in}
{\centering{\bf {Dr. G. Hemantha Kumar }} is currently the Chairman, Department of Studies in Computer Science, University of Mysore, Mysore, India. His Qualifications include B.Sc, B.Ed, M.Sc, Ph.D.}\\\\
\end{minipage}
He was awarded Ph.D. in Computer Science from University of Mysore. He has over 200 publications in all leading international and national journals as well as conferences.His current research interest includes
Numerical Techniques, Digital Image Processing, Pattern Recognition and Multimodal
Biometrics.\\\\
\noindent{\includegraphics[width=1in,height=1.7in,clip,keepaspectratio]{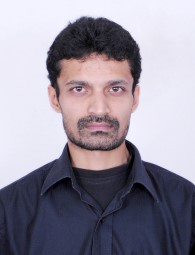}}
\begin{minipage}[b][1in][c]{2.0in}
{\centering{\bf{Manoj Krishnaswamy }}is a Research Scholar, Department of Studies in Computer Science, University of Mysore, Mysore, India. 
His Qualifications include B.E. in CompSci from R.V.C.E (B.U.) and M.Tech. in CompSci from M.V.J.C.E (V.T.U.)}\\\\
\end{minipage}
His current research interest includes Image Processing, Biometrics and Template Security.\\\\


\begin{thebibliography}{00}

\bibitem{1}
{N. K. Ratha, J. H. Connell, and R. M. Bolle}, ``{Enhancing security and privacy in biometrics-based authentication systems}," \emph{IBM systems Journal}, vol. 40, pp. 614-634, 2001.
\bibitem{2}
{S. Tulyakov, F. Farooq, and V. Govindaraju}, ``{Symmetric Hash Functions for Fingerprint Minutiae}," \emph{Proc. Int'l Workshop Pattern Recognition for Crime Prevention, Security, and Surveillance}, pp. 30-38, 2005.
\bibitem{3}
{A. B. J. Teoh, A. Goh, and D. C. L. Ngo}, ``{Random Multispace Quantization as an Analytic Mechanism for BioHashing of Biometric and Random Identity Inputs}," \emph{Pattern Analysis and Machine Intelligence, IEEE Transactions on}, vol. 28, pp. 1892-1901, 2006.
\bibitem{4}
{M. Savvides, B. V. K. V. Kumar, and P. K. Khosla}, ````{Corefaces"- Robust Shift Invariant PCA based Correlation Filter for Illumination Tolerant Face Recognition}," presented at \emph{IEEE Computer Society Conference on Computer Vision and Pattern Recognition (CVPR'04)}, 2004.
\bibitem{5}
{M. A. Dabbah, W. L. Woo, and S. S. Dlay}, ``{Secure Authentication for Face Recognition}," presented at \emph{Computational Intelligence in Image and Signal Processing}, CIISP 2007, IEEE Symposium on 2007.
\bibitem{6}
{I. Sobel}, ``{Neighborhood coding of binary images for fast contour following and general array binary processing}," \emph{Compute, Graphics Image process}, pp. 127-135, 1987.
\bibitem{7}
{L. G. Roberts}, ``{Machine perception of threedimensional solids, In Optical and Electrooptical Information processing}," \emph{MIT Press, Cambridge, MA}, 1965.
\bibitem{8}
{W. Frei and C.-C. Chen}, ``{Fast boundary detection: A generalization and a new algorithm}." \emph{lEEE Trans. Comput.}, vol. C-26, no. 10, pp. 988-998, 1977.
\bibitem{9}
{J.M.S. Prewitt}, ``{Object enhancement and extraction, in: B.S. Lipkin, A. Rosenfeld (Eds.)}," \emph{Picture Analysis and Psychopictorics, Academic Press, New York}, 1970.
\bibitem{10}
{Dodis, Y., Ostrovsky, R., Reyzin, L., Smith, A.}, ``{Fuzzy extractors: How to generate strong keys from biometrics and other noisy data}," \emph{Tech. Rep. 235, Cryptology ePrint Archive}, 2006.
\bibitem{11}
{Hao, F., Anderson, R., Daugman, J.}, ``{Combining crypto with biometrics effectively}," \emph{IEEE Trans. Comput. 55 (9)}, 1081-1088, 2006.
\bibitem{12}
{Nandakumar, K., Jain, A.K., Pankanti, S.}, ``{Fingerprint-based fuzzy vault: Implementation and performance}," \emph{IEEE Trans. Inform. Forensics Security 2 (4)}, 744-757, 2007.
\bibitem{13}
{Sutcu, Y., Li, Q., Memon, N.}, ``{Protecting biometric templates with sketch: Theory and practice}," \emph{IEEE Trans. Inform. Forensics Security 2 (3)}, 503-512, 2007.
\bibitem{14}
{C. T. Hsu and J. L. Wu}, ``{Hidden Digital Watermarks in Images}," \emph{IEEE Trans. On Image Processing}, vol. 8, no. 1, pp. 58-68, Jan. 1999.
\bibitem{15}
{Manvjeet Kaur, Dr. Sanjeev Sofat and Deepak Saraswat}, ``{Template and Database Security in Biometrics Systems: A Challenging Task}," published in \emph{International Journal of Computer Applications}, (0975 - 8887), Volume 4 - No.5, July 2010.
\bibitem{16}
{G. Hemantha Kumar and Manoj Krishnaswamy}, ``{Cancellable Face Biometrics Using Image Blurring}," \emph{International Journal of Machine Intelligence}, Vol. 3 Issue 4/5, p272, 2011.
\bibitem{17}
{AT{\&}T Laboratories Cambridge}, ``{The Database of Faces},"
\emph{http://www.cl.cam.ac.uk/research/dtg/\\attarchive/facedatabase.html}
\bibitem{18}
{Yale Face Database}, \emph{http://cvc.yale.edu/projects/\\yalefaces/yalefaces.html}
\bibitem{19}
{Vidit Jain, Amitabha Mukherjee}, ``{The Indian Face Database}," 
\emph{http://vis-www.cs.umass.edu/$\sim $vidit/IndianFaceDatabase/}
\bibitem{20}
{"Announcing the ADVANCED ENCRYPTION STANDARD (AES)},"
\emph{Federal Information Processing Standards Publication 197, United States National Institute of Standards and Technology (NIST)}, November 26, 2001, Retrieved October 2, 2012.
\end{thebibliography}
\end{document}